\documentclass[10pt,twocolumn,letterpaper]{article}

\usepackage{cvpr}              

\usepackage{graphicx}
\usepackage{amsmath}
\usepackage{amssymb}
\usepackage{booktabs}

\usepackage{algorithm}
\usepackage{algpseudocode}

\newcommand{\Iref}{I^\textit{R}}
\newcommand{\Itar}{I^\textit{tar}}
\newcommand{\Iout}{I^\textit{*}}
\newcommand{\IR}{\textit{IR}}
\newcommand{\tar}{\textit{tar}}
\newcommand{\coarse}{\textit{coarse}}

\newcommand{\T}{\textup{T}}
\newcommand{\R}{\textit{R}}
\newcommand{\uP}{\textup{P}}
\newcommand{\uS}{\textup{S}}

\usepackage{xcolor}
\usepackage{mathtools}
\usepackage{soul}

\newif\ifdraft
 \draftfalse

\ifdraft
\newcommand{\dcc}[1]{{\color{red}[\textbf{Danny:} #1]}}
\newcommand{\dc}[1]{{\color{red} #1}}
\newcommand{\dlc}[1]{{\color{purple}[\textbf{DL:} #1]}}
\newcommand{\dl}[1]{{\color{purple}#1}}

\newcommand{\yang}[1]{{\color{blue} #1}}
\newcommand{\yzc}[1]{{\color{blue}[\textbf{YZ:} #1]}}

\newcommand{\drop}[1]{}

\else
\newcommand{\dcc}[1]{}
\newcommand{\kac}[1]{}
\newcommand{\avc}[1]{}
\newcommand{\cqc}[1]{}
\newcommand{\dc}[1]{{\color{black}#1}}

\newcommand{\dlc}[1]{}
\newcommand{\yzc}[1]{}
\newcommand{\dl}[1]{{\color{black}#1}}
\newcommand{\yang}[1]{{\color{black} #1}}
\fi

\makeatletter
\DeclareRobustCommand\onedot{\futurelet\@let@token\@onedot}
\def\@onedot{\ifx\@let@token.\else.\null\fi\xspace}
\DeclareMathAlphabet\mathbfcal{OMS}{cmsy}{b}{n}

\def\eg{\emph{e.g}\onedot}

\def\ie{\emph{i.e}\onedot}

\def\etal{\emph{et al}\onedot}
\makeatother

\makeatletter
\def\blfootnote{\xdef\@thefnmark{}\@footnotetext}
\makeatother

\newif\ifwatermark
\watermarktrue
\draftfalse
\usepackage[pagebackref,breaklinks,colorlinks]{hyperref}

\usepackage[capitalize]{cleveref}
\crefname{section}{Sec.}{Secs.}
\Crefname{section}{Section}{Sections}
\Crefname{table}{Table}{Tables}
\crefname{table}{Tab.}{Tabs.}
\Crefname{figure}{Fig.}{Figs.}

\def\cvprPaperID{6971} 
\def\confName{CVPR}
\def\confYear{2024}

\begin{document}

\title{Generating Non-Stationary Textures using Self-Rectification}

\author{Yang Zhou\textsuperscript{\rm 1} \qquad Rongjun Xiao\textsuperscript{\rm 1} \qquad Dani Lischinski\textsuperscript{\rm 2} \qquad Daniel Cohen-Or\textsuperscript{\rm 3} \qquad Hui Huang\textsuperscript{\rm 1}\thanks{Corresponding author.} \\
\textsuperscript{\rm 1}Shenzhen University \qquad \textsuperscript{\rm 2}The Hebrew University of Jerusalem \qquad \textsuperscript{\rm 3}Tel Aviv University \\ 
}

\maketitle

\begin{abstract}
This paper addresses the challenge of example-based non-stationary texture synthesis. We introduce a novel two-step approach wherein users first modify a reference texture using standard image editing tools, yielding an initial rough target for the synthesis. Subsequently, our proposed method, termed ``self-rectification", automatically refines this target into a coherent, seamless texture, while faithfully preserving the distinct visual characteristics of the reference exemplar.
Our method leverages a pre-trained diffusion network, and uses self-attention mechanisms, to gradually align the synthesized texture with the reference, ensuring the retention of the structures in the provided target. Through experimental validation, our approach exhibits exceptional proficiency in handling non-stationary textures, demonstrating significant advancements in texture synthesis when compared to existing state-of-the-art techniques.
Code is available at \url{https://github.com/xiaorongjun000/Self-Rectification}
\end{abstract}

\section{Introduction}

Example-based texture synthesis aims to generate a texture that faithfully captures all the visual characteristics of a provided reference texture exemplar. The key challenge is to generate a texture that visually mimics the reference, while avoiding exact replication and without producing conspicuous, unnatural artifacts.

Over the past decades, numerous methods have emerged for synthesizing textures from examples, and many have demonstrated impressive results, particularly for homogeneous textures that can be accurately captured by stationary models \cite{Wei09}. Nonetheless, a significant challenge remains when dealing with real-world textures that are inherently inhomogeneous, and non-stationary.

\begin{figure}[t]
 \begin{center}
    \centering
    \captionsetup{type=figure}
    \includegraphics[width=\columnwidth]{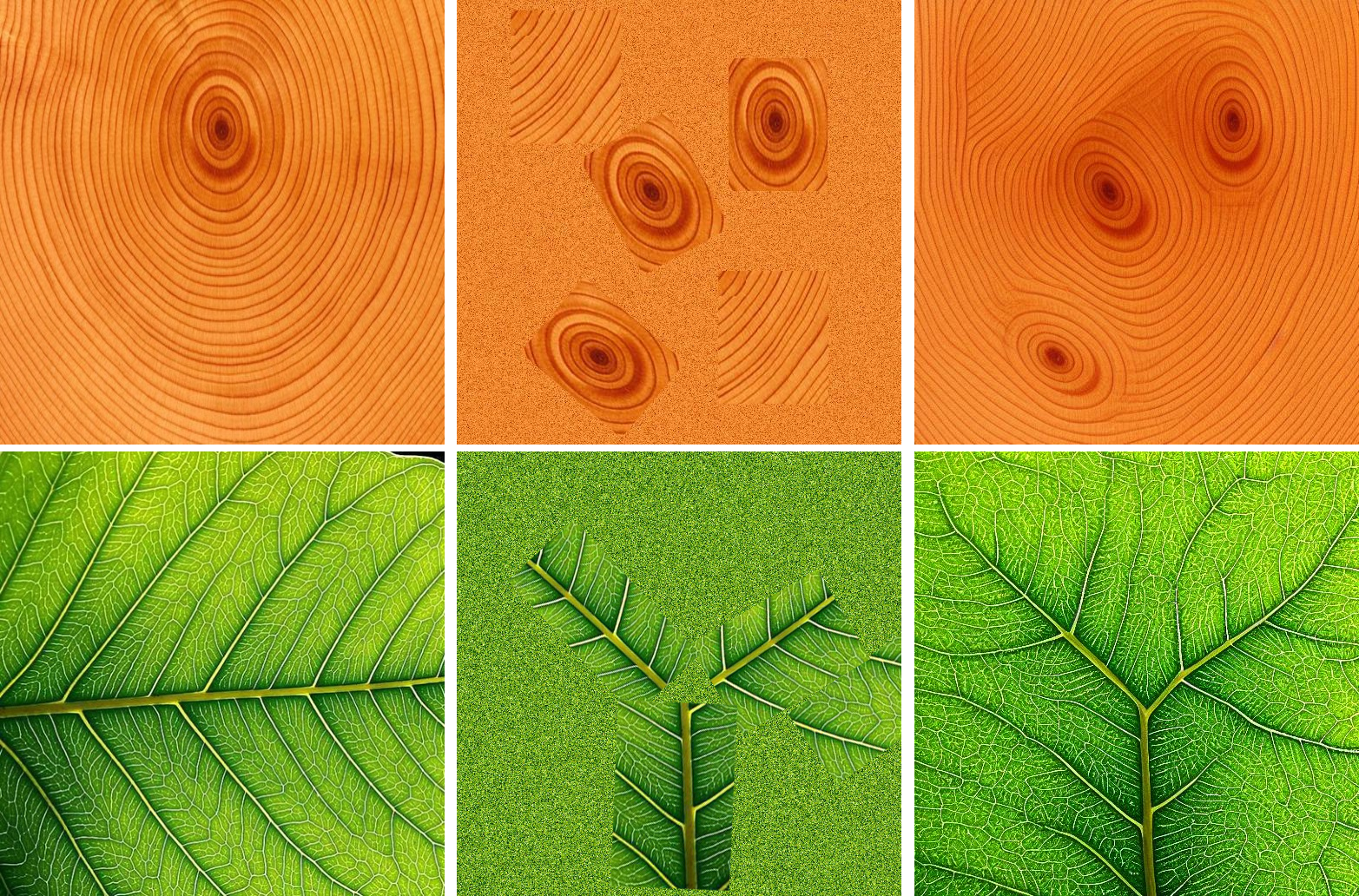}
    \caption{Our method takes as input a reference texture (left), and a crude target texture provided by the user (middle column), which may lack coherence and completeness. Self-rectification is used to transform the target into a visually coherent texture (right) that complies with the structure of the crude target, while exhibiting the visual characteristics of the reference texture.}
    \label{fig:teaser}
\end{center}
\end{figure}

Non-stationary textures exhibit distinctive attributes, such as sprawling irregular large-scale structures or variations in attributes such as color, local orientation, and local scale. \yang{\Cref{fig:teaser} (left) shows two examples.} Mimicking such complex structures and variations through example-based synthesis is a long-standing challenging task~\cite{Rosenberger:2009}. 

The emergence of neural networks has provided powerful means to deal with non-stationary textures. Zhou~\etal~\cite{TexExp} introduced a method that involves overfitting a GAN, where the encoder extracts structural guidance for the decoder. This approach provides a viable means of spatially extending non-stationary textures while preserving the visual characteristics of the reference.  However, this technique requires an extremely long optimization process for a single texture examplar, and moreover, it fails to provide any controllability or editability.

In this paper, we introduce a two-step lazy-editing technique where the user first edits the given reference texture using conventional image editing tools, to obtain an extremely rough initial result, which may be incomplete and incoherent, as demonstrated in \Cref{fig:teaser} (middle). Next, our technique automatically rectifies the edited texture into a regularized, coherent and seamless texture that follows the rough target, while retaining the local characteristics of the reference; see, \eg, \Cref{fig:teaser} (right). We term this regularization process \textit{self-rectification}, since the texture is rectified based on the input texture itself. 

\dl{
To self-rectify the crude target, we use a pre-trained diffusion network, and synthesize the final rectified result while utilizing the intermediate byproducts obtained by inverting both the initial target and the reference exemplar.
Inspired by editing approaches that leverage self-attention~\cite{masactrl, infusion, alaluf2023crossimage}, we extract certain self-attention features of the diffusion model during the inversion of the rough target, as well as the reference. These features are then injected into various steps in the inversion and denoising of the generated texture.
Such feature injection forms ``cross-attention'' between the reference and target features~\cite{masactrl,alaluf2023crossimage}, and thereby, in the course of the diffusion process, the target texture is progressively refined to be increasingly \emph{locally} similar to the reference, while retaining the global structure prescribed by the provided target. This self-rectification occurs in two passes, first addressing the larger scale structure and subsequently focusing on finer local details.
}

To \dc{enhance} the synthesis quality, we augment the reference texture with several transformed copies of it, \dl{which increases the diversity of the reference patterns, thereby improving the compatibility between the reference and the synthesized result.}  The augmented source features are injected into the corresponding target layers as well. \dc{Furthermore, we show that } our method can \dc{also be} applied to lazy editing of natural images, \dc{using the same means} of synthesizing highly non-stationary textures.  Experiments show that our method can deal with a large scope of non-stationary textures, with unprecedented \dc{flexibility} and quality, compared to the state-of-the-art.

\section{Related Work} \label{sec:related}

\paragraph{Non-stationary texture synthesis.} 
Early methods in example-based texture synthesis mainly focused on synthesizing textures with stationary characteristics~\cite{Efros99, Wei2000, Efros:2001,kwatra2003, kwatra2005, Wexler07, Self-tuning}. To cope with non-stationary textures, researchers typically involve additional guidance for the control of distinctive attributes, \eg, using label maps to guide the layout synthesis of composite textures~\cite{Hertzmann01,Rosenberger:2009,lockerman16}, representing the spatial variations of weathering textures with age/progression maps~\cite{Wang2006,Bellini2016}, and describing the local orientations of directional textures with vector fields~\cite{PG15, EG17}. Although the aforementioned methods have shown success, the guidance required from the user is tedious to provide, yet still limited. 

In the deep learning era, neural methods for texture generation have rapidly emerged, either through new texture losses for texture optimization~\cite{Gatys15, Heitz_2021_CVPR, Zhou2023GCD} or via training generative networks~\cite{Ulyanov2016, Li2016, Jetchev2016, Bergmann2017, Sendik2017, tilegan2019,fastshi2020,structuralbenaim2021, SeamlessGAN2022, TexExp, InGAN, SinGAN}. 
Among these methods, Sendik and Cohen-Or~\cite{Sendik2017} attempted to preserve the non-local structures of non-stationary textures by regularizing the feature correlation between different locations. Zhou \etal~\cite{TexExp} proposed overfitting a GAN to learn the expansion from a small texture block to a larger one containing it. Their approach involves the GAN's encoder extracting the global structure of an input texture, duplicated by bottleneck residual blocks before decoding. Although these trained models can effectively extend non-stationary textures, their overfitting nature severely restricts their generalization and controllability.

\paragraph{Diffusion-based image synthesis.}
The emergence of large-scale generative diffusion models, such as Stable Diffusion (SD)~\cite{stablediffusion} and DALLE-2~\cite{ramesh2022hierarchical}, has revolutionized image synthesis due to their unprecedented generation power.
%
To make the pre-trained diffusion models synthesize on image conditions, many solutions are proposed, including optimizing prompt tokens~\cite{gal2022textual, alaluf2023neural}, fine-tuning the entire model~\cite{ruiz2022dreambooth}, or training an additional adapter~\cite{yang2022paint, ye2023ip-adapter, ruiz2023hyperdreambooth,hu2021LoRA,zhang2023ControlNet}. 

Recently, researchers have investigated the role of the intermediate attention maps and features in the diffuser~\cite{hertz2022prompt, chefer2023attendandexcite, parmar2023zero, Tumanyan_2023_pnp}, finding them crucial for layout/structure synthesis.
The method we present in this paper, builds upon the mechanism of injecting keys and values from attention layers of one diffusion process into another, as a means for transfering visual features between images~\cite{masactrl,infusion,alaluf2023crossimage}. We use such injection to ``copy'' the local patterns from a source texture to a target one. 
\dl{
In addition to the injection we also incorporate coarse-to-fine and data augmentation schemes, forming the self-rectification framework for generating non-stationary textures.
}

\begin{figure}[t]
	\centering
	\includegraphics[width=1.0\linewidth]{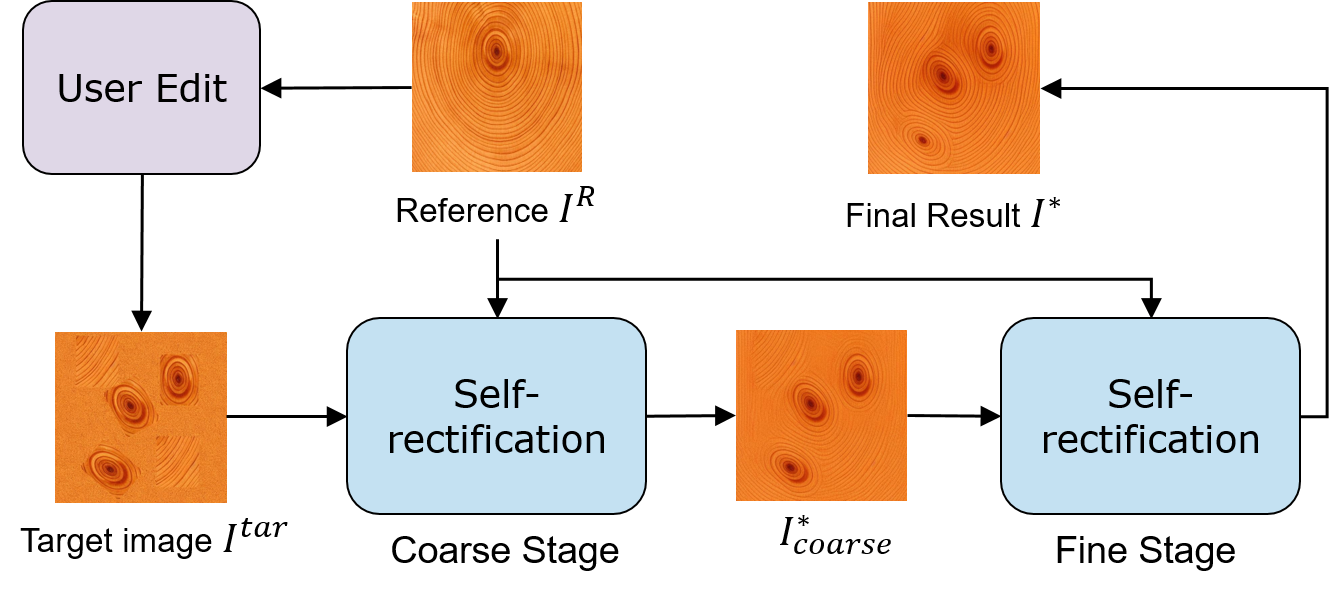}
        \caption{Framework overview. Given a reference texture $I^\R$, we allow the user to quickly build a target image $\Itar$ in a lazy-editing manner. A coarse-to-fine synthesis is performed by running self-rectification twice. The coarse stage synthesizes a coarse yet complete overall structure, and the fine stage refines its output $I_\coarse^*$ with finer and more accurate details, producing the final result $I^*$.}
	\label{fig:overview}
\end{figure}

\begin{figure*}[t]
	\centering
	\includegraphics[width=1.0\linewidth]{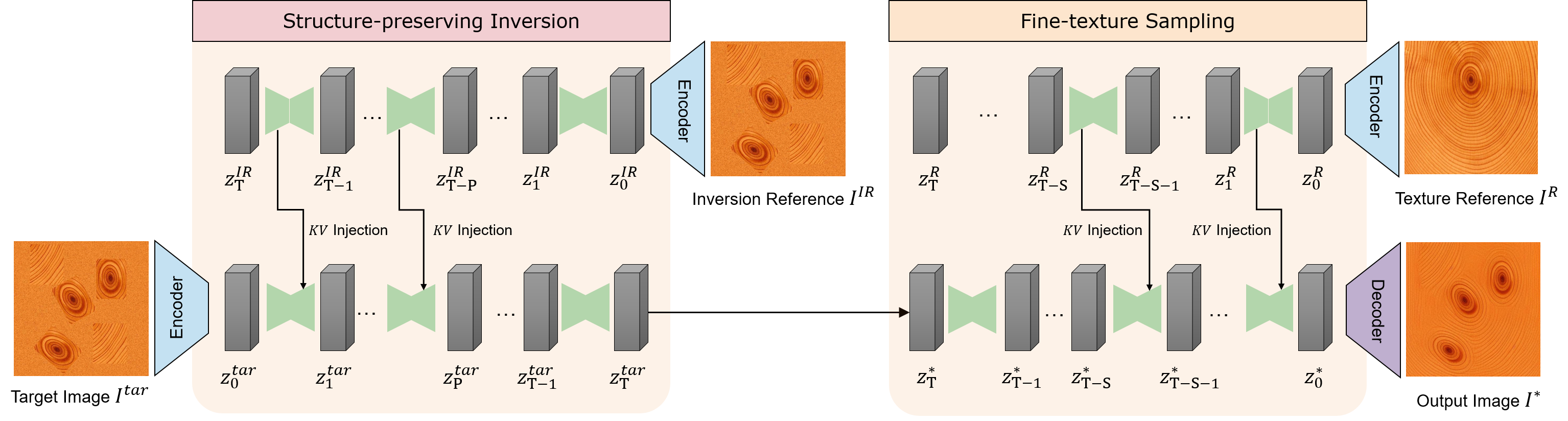}
	\caption{Our self-rectification synthesizes an output texture $I^*$ via structure-preserving inversion from a rough target image $\Itar$ and fine texture sampling using the reference $\Iref$. Both processes require the injection of self-attention features ($KV$) from the DDIM inversion of a corresponding reference. More specifically, for structure-preserving inversion, the reference is the target image itself, denoted as $I^\IR$. For fine texture sampling, the input exemplar $\Iref$ is used to inject features that help to synthesize a plausible output with fine texture details. }
	\label{fig:target-refine}
\end{figure*}

\section{Method}

Our method employs a two-step process for synthesizing a new texture $I^*$ resembling a provided non-stationary reference texture $\Iref$, 
while guided by a user-provided rough target $\Itar$, as depicted in \Cref{fig:overview}.
Initially, the user rapidly creates $\Itar$, by assembling patches from the source reference texture. Such a lazy-edit forms a basic layout for the output image, $\Iout$.
However, this initial sketch may be incoherent or incomplete, 
hence necessitating rectification. Subsequently, the texture undergoes self-rectification to align with both the user's coarse layout and the reference image's detailed texture characteristics. This self-rectification process is depicted in \Cref{fig:overview} and is executed in two stages: coarse rectification, followed by a finer one.

Considering that the rough target $\Itar$ comprises patches derived from the reference texture $\Iref$, we implement a process of self-rectification, detailed below. In a broad sense, we utilize a pre-trained latent diffusion model to invert both the target image and the reference image into an initial latent noise. This is followed by employing feature injection during the sampling process of $\Iout$. The features from $\Itar$ guide the structural synthesis and the features from $\Iref$ guide the fine texture details.

Below, we first briefly review diffusion models and their self-attention and the cross-KV-injection mechanisms, and then proceed to present our self-rectification technique.

\subsection{DDIM Sampling and Inversion}\label{sec:Preliminaries}

Denoising diffusion models~\cite{ho2020ddpm, song2020ddim} involve two processes: a noising process that gradually transforms an input image into Gaussian noise, and a denoising/sampling process that generates images from Gaussian samples. Using \emph{DDIM sampling}~\cite{song2020ddim}, starting from a noise sample $z_\T$, one can generate a clean sample $z_0$ via $\T$ deterministic steps:
\begin{equation}
\label{eq:sampling}
z_{t-1}=\sqrt{\bar{\alpha}_{t-1}} f_{\theta}\left(z_{t}, t\right)+\sqrt{1-\bar{\alpha}_{t-1}} \epsilon_{\theta}\left(z_{t}, t\right),
\end{equation}
where $\epsilon_{\theta}$ is a noise prediction network conditioned on the current noisy sample $z_t$ and timestamp $t$. $\bar{\alpha}_t$ is the noise scaling factor defined in~\cite{song2020ddim}, and $f_{\theta}\left(z_{t}, t\right)$ is
\begin{equation}
f_{\theta}\left(z_{t},t\right)=\frac{z_t-\sqrt{1-\bar{\alpha}_t} \epsilon_{\theta} \left(z_t,t \right) }{\sqrt{\bar{\alpha}_t} } .
\end{equation}

In the opposite direction, \emph{DDIM inversion}~\cite{song2020ddim} of a given image $z_0$ is the process of incrementally adding deterministic noise, until obtaining $z_\T$:
\begin{equation}
\label{eq:inversion}
z_{t+1}=\sqrt{\bar{\alpha}_{t+1}} f_{\theta}\left(z_{t}, t\right)+\sqrt{1-\bar{\alpha}_{t+1}} \epsilon_{\theta}\left(z_{t}, t\right).
\end{equation}
Such inversion leads to a nearly faithful reconstruction~\cite{song2020ddim}.

\subsection{Stable Diffusion and Self-Attention}

We base our texture synthesis framework on Stable Diffusion (SD)~\cite{stablediffusion}, which is a pretrained latent diffusion model consisting of an encoder $\mathcal{E}$ that maps an input image into the latent space, and a decoder $\mathcal{D}$ that reconstructs a latent code back into image space. Both DDIM sampling and DDIM inversion are performed in the SD latent space.

The noise predictor ${\epsilon}_{\theta}$ of SD is a large-scale U-Net that contains multiple self-attention modules~\cite{vaswani2017attention}. Each self-attention layer transforms its input intermediate feature map (also called spatial features) into an attended representation by the following equation:
\begin{equation}
\label{eq:attention}
\textup{Att}\left(Q,K,V\right)=\textup{Softmax}\left(\frac{QK^T}{\sqrt{d}}\right)V,
\end{equation}
where $Q, K$, and $V$ are the queries, keys, and values, respectively, obtained by learned linear projections of the same input spatial features, having dimension $d$. The self-attention mechanism uses the similarities between the queries and keys as attention scores to weigh the importance or relevance of the values. 
Relevant info is thus aggregated as the attended representation. 

\subsection{KV-Injection}

The self-attention features contain rich information about both the large-scale structures and local fine details of an input image~\cite{Tumanyan_2023_pnp, masactrl, alaluf2023crossimage}. 
Tumanyan \etal~\cite{Tumanyan_2023_pnp} demonstrate that by injecting the spatial features and the queries and keys ($QK$) from the self-attention layers of a source (guidance) image into the corresponding layer of the generated target image during the sampling process, one can preserve the layout of the source while modifying its appearance. Injecting only the $KV$ features of a source reference, instead, transfers its appearance, including textures, to the generated target~\cite{masactrl, alaluf2023crossimage}. 

In our case, the target image is a rough and incomplete guidance, typically containing only a few source patches rotated and placed freely by the user. A complete overall structure needs to be synthesized reasonably yet conforming to user's constraints. Locally, however, the resulting texture should resemble the reference. To achieve these goals, we adapt the $KV$-injection into both the DDIM inversion and sampling, bringing a novel self-rectification operation.


\begin{figure}[t]
	\centering
	\includegraphics[width=1.0\linewidth]{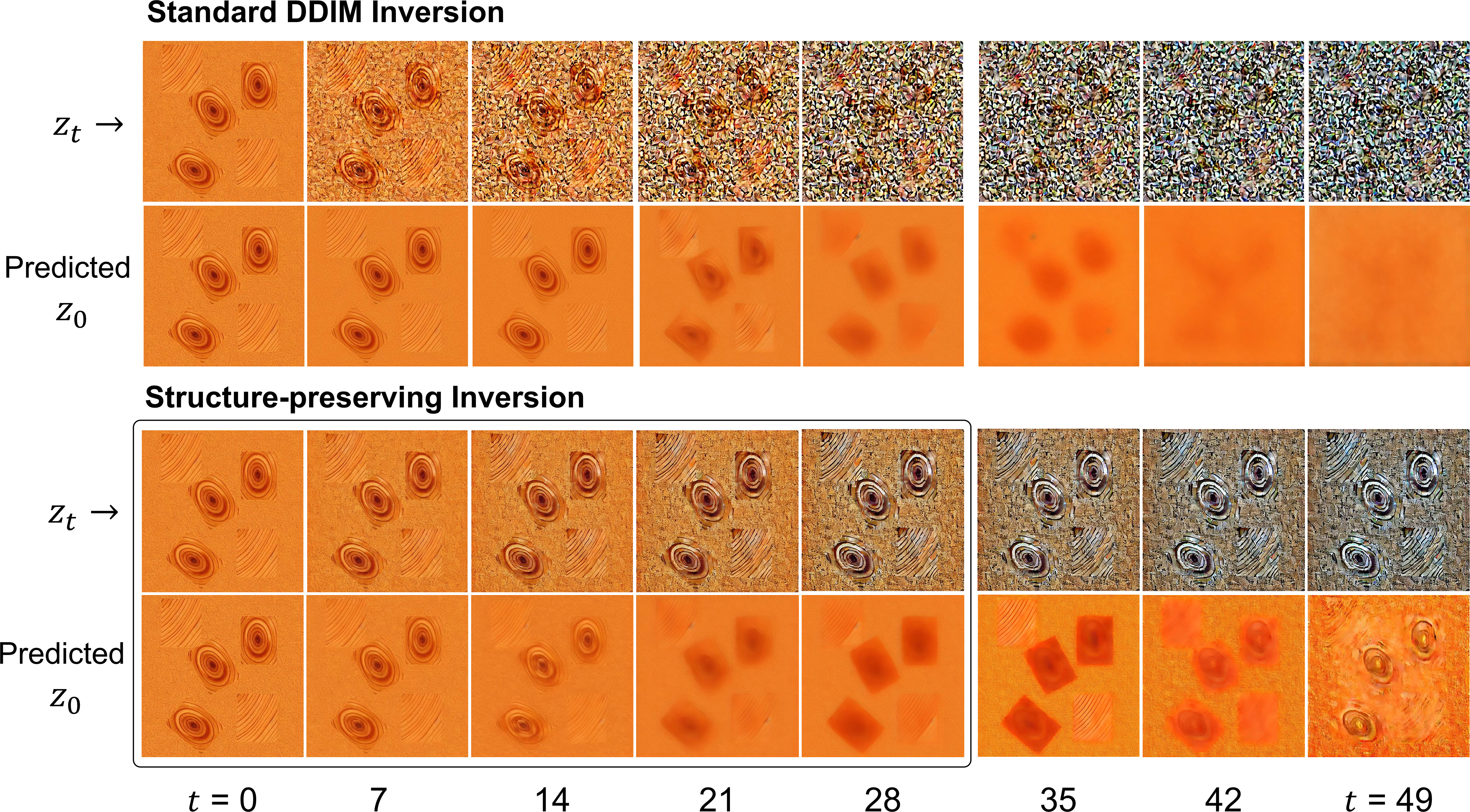}
	\caption{Visualization of the intermediate latent codes in the inversion. For the standard DDIM inversion (top), the U-Net predicts noise to diffuse the distinctive structures so as to transform the input into Gaussian noise. \yang{In contrast, our structure-preserving inversion (bottom) reserves the distinctive patterns from user edits along the inversion process. See texts in \cref{subsec:refinement} for more details.} }
	\label{fig:latent_vis}
\end{figure}

\subsection{Self-rectification}
\label{subsec:refinement}

As shown in \Cref{fig:target-refine}, our self-rectification consists of two parts: structure-preserving inversion and fine texture sampling. Both are performed in the SD latent space. The last latent code of inversion is used as the starting code of the sampling. The core modification to both processes is the $KV$-injection of self-attention features.

\paragraph{Structure-preserving inversion.}

The standard DDIM inversion (Eq.~\eqref{eq:inversion}) progressively transforms an input image (SD latent code)  into pure Gaussian noise. At each time step, the noise to be added is predicted by the U-Net. As visualized in \Cref{fig:latent_vis}, for a given target image, the noise predicted by the U-Net for early time steps (\eg, $t\leq20$) is mainly distributed to ``diffuse'' the prominent structures so that the distinctive patterns from user edits get scattered and random. \dl{As the inversion progresses, the magnitude of the noise added in each step becomes smaller and similar for all latent pixels,} since the ``diffusion'' is getting close to done. 
As no text prompt is involved, the noise prediction in each inversion step is dominantly determined by the self-attention mechanism in the U-Net. Our key observation is that if we inject the $KV$ features from a large time step $t_1$ ($\gg\T/2$) into an early time step $t_2$ ($\ll\T/2$), the noise predicted at $t_2$ will be smaller and more spatially uniform. The distinctive patterns of $\Itar$, reflecting the user's edits, are thus better preserved. Hence, we refer to this process as \emph{structure-preserving inversion}.
%

Specifically, given a rough target image $\Itar$, we invert it twice. The first inversion is a standard DDIM inversion. The produced self-attention features along the noising steps are regarded as the \emph{inversion reference} (IR) for the second inversion. The self-attention during the second inversion, at time step $t$, is now given by:
\begin{equation}
\label{eq:structure-preserving}
\textup{Att}\left(Q_t^{\tar},K_{\T-t}^\textit{IR},V_{\T-t}^\IR\right)=\textup{Softmax}\left(\frac{Q_t^{\tar}\left(K_{\T-t}^\IR\right)^T}{\sqrt{d}}\right)V_{\T-t}^\IR,
\end{equation}
where $\T$ denotes the total number of steps \dl{(we use $\T$ = 50).} Here $KV$ features are injected in a reverse order in the second round. We would like to stress that we have also experimented with using an offset, \ie, replacing $\T-t$ with $t+\textit{offset}$ in the above equation, but found no advantage (see supplementary for more analysis and comparisons).

During the second inversion, we do not perform $KV$ injections for all time steps. We set a parameter $\uP$ that defines the time step beyond which we continue with standard inversion. \Cref{fig:latent_vis} shows the effect of our structure-preserving inversion, where $\uP=30$. Since the final latent code after the second inversion still contains rich information about the distinctive structures from $\Itar$, the subsequent sampling that starts from this latent code is guided to synthesize a global structure that complies with $\Itar$.

\begin{figure}[t]
	\centering
	\includegraphics[width=1.0\linewidth]{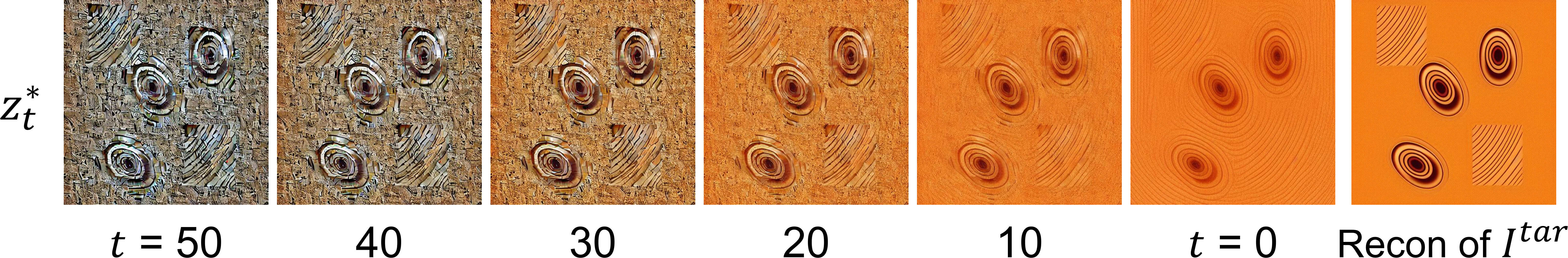}
	\caption{Visualization of the intermediate latent codes in the fine texture sampling. Here, for the first 20 steps (from $t=50$ to $30$), we perform the standard DDIM sampling to reconstruct the target layout. Next, we perform $KV$-injection in the remaining sampling steps ($t=30$ to 0), to synthesize fine textures for the output image. \yang{The rightmost shows the result produced by simply performing standard DDIM sampling for all steps, \ie, $\uS=50$. No additional structure is synthesized to complete the user edits.}} 
	\label{fig:latent_vis_sampling}
\end{figure}


\paragraph{Fine texture sampling.}

We start the DDIM sampling from the final structure-preserving latent code. The sampling uses the U-Net to predict the noise to be removed, as defined in Eq.~\eqref{eq:sampling}. 
However, simply performing standard DDIM sampling for all denoising steps results in a nearly reconstructed target image since  DDIM sampling is deterministic.
Therefore, we set the first $\uS$ steps (\ie, from time step $\T$ to $\T-\uS$) to reconstruct the target layout to a certain extent. For the remaining $\T-\uS$ denoising steps, we synthesize the output image $I^*$ by matching fine textures from the reference via $KV$-injection from the reference $\Iref$. 

To this end, we first DDIM-invert the reference texture $\Iref$. At denoising time step $t$ \yang{($t>\T-\uS$)}, the $KV$ features extracted during the inversion are injected into the corresponding self-attention layers of the synthesized texture:
\begin{equation}
\label{eq:texture-matching}
\textup{Att}\left(Q_t^{*},K_{t}^{\R},V_{t}^{\R}\right)=\textup{Softmax}\left(\frac{Q_t^{*}\left(K_{t}^{\R}\right)^T}{\sqrt{d}}\right)V_{t}^{\R}.
\end{equation}
This forms a cross-image attention, where corresponding fine local patterns in the reference are transferred to the output image in a plausible manner. \Cref{fig:latent_vis_sampling} visualizes the intermediate process of our texture sampling, where $\uS = 20$.

\subsection{Implementation details and data augmentation}

In the user editing phase, we fill the background of the target canvas with pixels randomly drawn from the source texture, such that the encoding of the target image would not deviate too far from the source in the SD latent space. 
As the self-rectification is performed twice, the output of coarse stage $I^*_\coarse$, will be used as the input target image of fine stage to produce the final output $I^*$. 
In contrast, for the inversion reference $I^\IR$ involved in structure-preserving inversion of the two self-rectifications, we use the same initial target image that contains the original user edits. See the supplementary for the full algorithm pseudo-code.

Considering the parameter settings: since the coarse stage aims for structure synthesis, relatively large $\uP$ and $\uS$ are required in self-rectification, and vice versa in the fine stage. More specifically, let $\uP_1$, $\uP_2$, $\uS_1$, and $\uS_2$, denote the parameters used in the two rounds of structure-preserving inversion and fine texture sampling. We typically set $\uP_1=20, \uP_2=5, \uS_1=20,$ and $\uS_2=5$.
Following~\cite{masactrl}, we choose the $KV$ features from the 10th to 15th self-attention layers of the U-Net decoder part. 

To further improve the synthesis quality, especially when dealing with textures containing a dominant directional structure, such as the leaf shown in \Cref{fig:teaser}, we can introduce a few transformed images (flips and rotations) to augment the reference texture. We concatenate the new attention features from the augmentation to the original reference feature. For example, when we have $n$ augmented source textures, the $K^R$ and $V^R$ in Eq.~\eqref{eq:texture-matching} is now given by
\begin{equation}
\label{eq:augmentation}
\begin{cases}
K^{R} & =\textup{Concat}\left( K_{(0)}^R, K_{(1)}^R, \ldots,K_{(n)}^R \right)\\
V^{R} & =\textup{Concat}\left( V_{(0)}^R, V_{(1)}^R, \ldots,V_{(n)}^R \right)
\end{cases},
\end{equation}
where $ K_{(i)}^R,$ and  $V_{(i)}^R$ are the self-attention features acquired from the DDIM inversion of the augmented references.
\section{Experiments} \label{sec:results}

We apply our method with Stable Diffusion with publicly available checkpoints v1.4. All experiments were conducted on a single Quadro P6000 24G GPU. The inference time synthesizing an image of 512$\times$512 pixels takes about three minutes for our coarse-to-fine self-rectification.

\begin{figure*}[htpb]
	\begin{center}
		\centering
		\captionsetup{type=figure}
		\includegraphics[width=\linewidth]{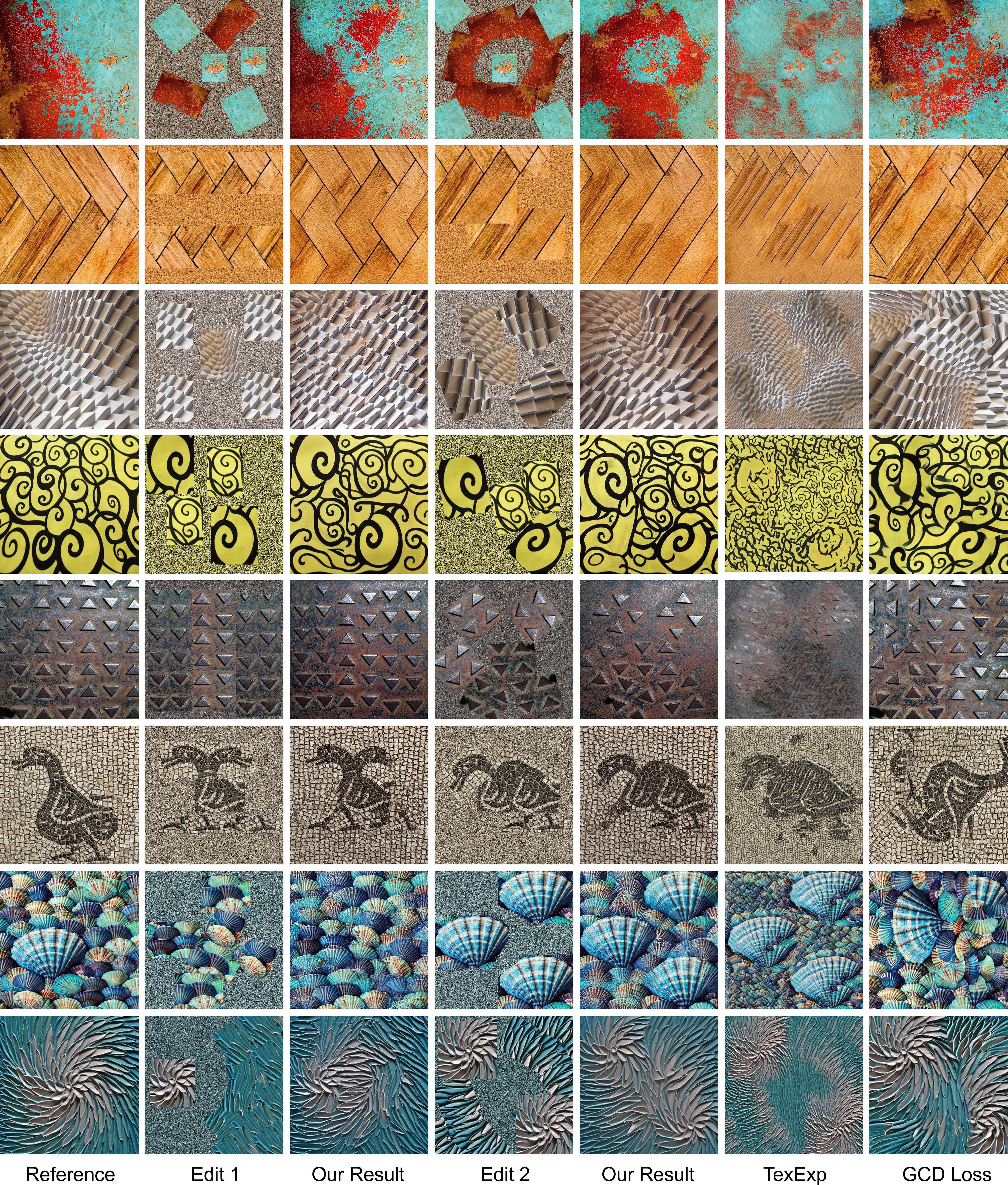}
		\caption{We sketch two targets for each reference, denoted as Edit 1 and Edit 2, respectively. Our self-rectification method generates textures with global structures that faithfully respect the user edits of the target images, while still producing high-quality texture details. In contrast, adversarial expansion (TexExp)~\cite{TexExp} does not capture the fine details well. Optimization by GCD Loss~\cite{Zhou2023GCD} reproduces the local textures, but does not always conform to the target image. Note that Edit 2 is used as the input for TexExp, and as the initialization in texture optimization of GCD Loss. More results are included in the supplementary.
		}
		\label{fig:gallery}
	\end{center}
\end{figure*}

\subsection{Evaluations and Comparisons}

We evaluate our method with non-stationary textures released by~\cite{TexExp}. Each example is resized to 512$\times$512 pixels as the reference, and we quickly built several different target images for each \yang{in PhotoShop} with \dc{just} a few lazy edits. Fig.~\ref{fig:gallery} shows a gallery of results generated by our method. As can be seen our method faithfully reproduces the delicate textures of the exemplar, their global structure and yet respecting the sparse edits of the target image. More results are included in the supplementary.

To compare with state-of-the-art texture synthesis methods, we fed the target images we sketched to the models trained by adversarial expansion (TexExp)~\cite{TexExp}. As can be seen in \Cref{fig:gallery}, TexExp failed to reproduce the fine textures of the reference, as the edited target images are unseen data to its training. We also tested the texture optimization based on a recently proposed textural loss (GCD Loss)~\cite{Zhou2023GCD}, where the target images are down-scaled as the initialization in its multi-resolution synthesis. Although plausible local textures are synthesized, the output of this method does not always conform to the user-edited layout (see \Cref{fig:gallery}).

\begin{figure}[t]
	\begin{center}
		\centering
		\captionsetup{type=figure}
		\includegraphics[width=\linewidth]{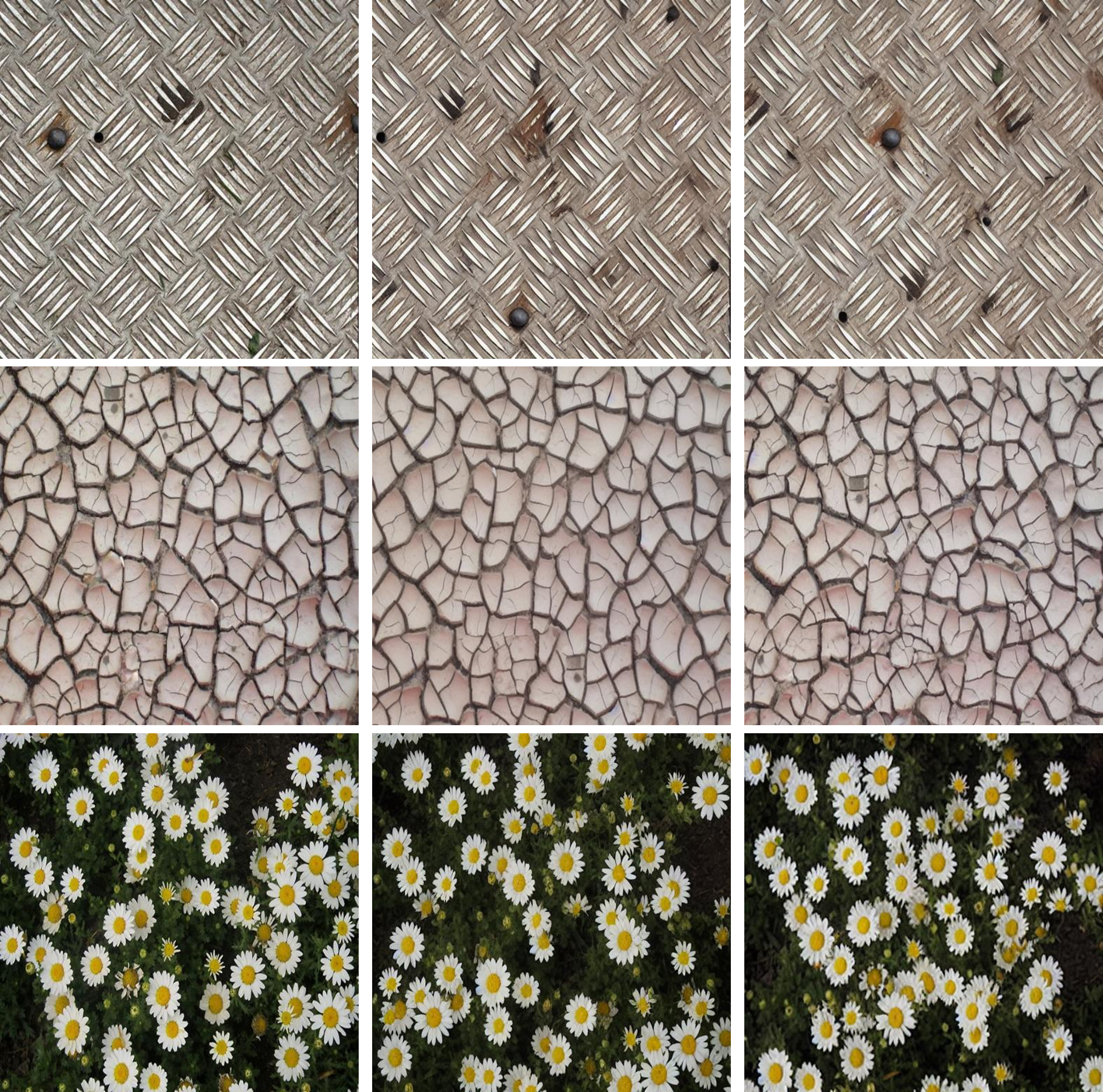}
		\caption{\yang{For nearly homogeneous textures (left), we can use patch shuffle of the reference to define a random target layout, where the shuffling could be performed in image space before \dc{applying} self-rectification (middle), or in latent space after the inversion in the first self-rectification (right).  }}
 
		\label{fig:stationary}
	\end{center}
\end{figure}

\dc{We have also applied our method on nearly homogeneous textures of stationary statistics. Since such textures do not have a prominent global structure, we can simply reshuffle patches of the reference to serve as the target. This, however, might cut some local elements in the example.
Alternatively, we can use the reference image as the target and shuffle its inversion code before the sampling, which better preserves the local texture elements. \Cref{fig:stationary} shows a few examples, demonstrating promising quality.} 

\begin{figure}[t]
	\begin{center}
		\centering
		\captionsetup{type=figure}
		\includegraphics[width=\linewidth]{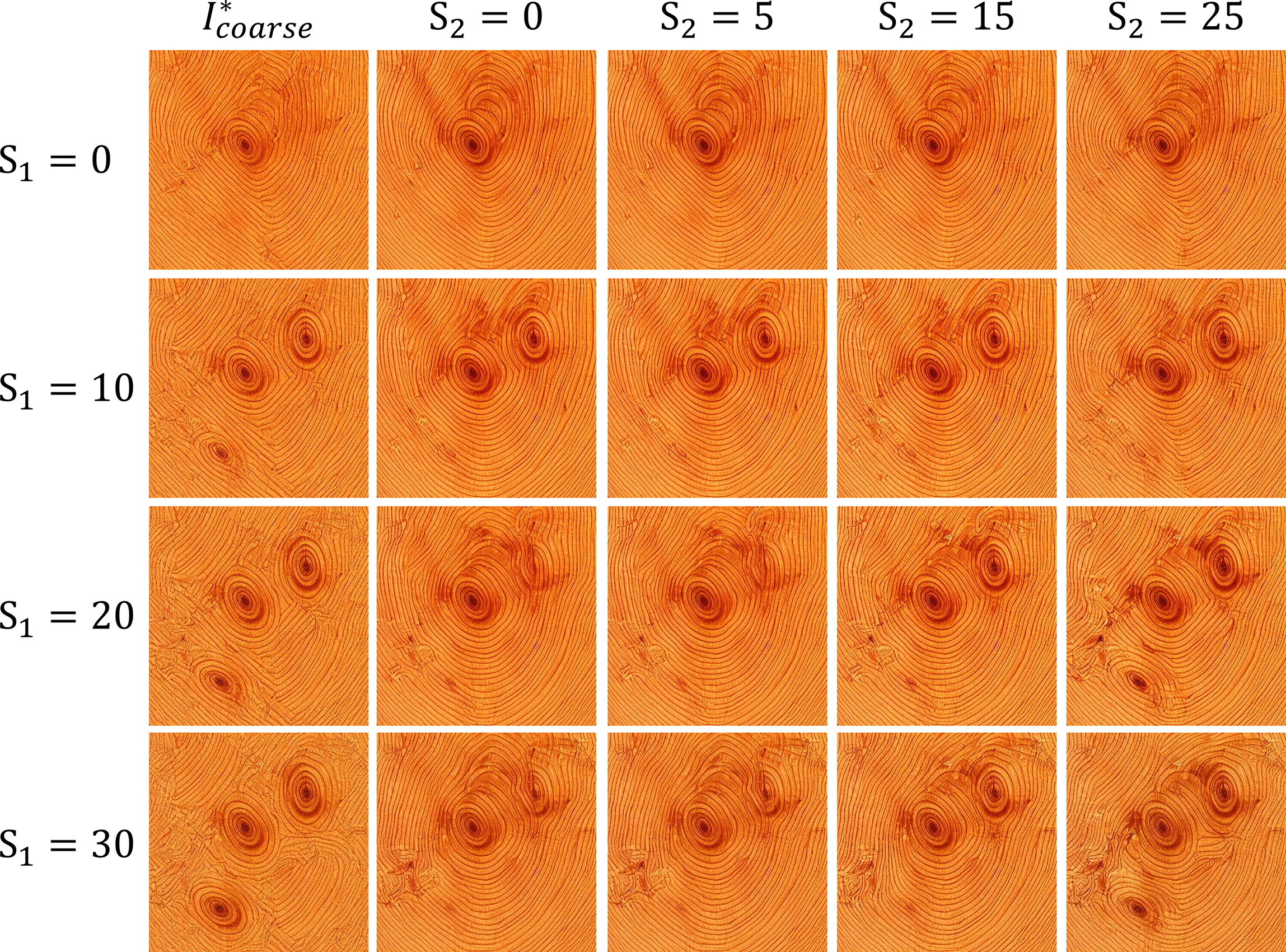}
		\caption{\yang{Ablation study of $KV$-Injection in texture sampling. We explore the parameter space for $\uS_1$ and $\uS_2$, with $\uP_1=\uP_2=0$, and adopt the wood texture and its edited target shown in Fig.~\ref{fig:teaser} for this test.}  
			The coarse self-rectification results are shown in the 1st column. 
			We can see that changing the starting time-step of $KV$-Injection in the sampling greatly affects the final output. However, none of these settings yields a qualified balance when considering both the target layout and the texture details. \yang{Note setting $\uS=50$ (\ie, no $KV$-Injection in sampling) only results in an approximate layout reconstruction without any details synthesized} (see Fig.~\ref{fig:latent_vis_sampling}).  } 
		\label{fig:s1s2}
	\end{center}
\end{figure}

\begin{figure}[t]
	\begin{center}
		\centering
		\captionsetup{type=figure}
		\includegraphics[width=\linewidth]{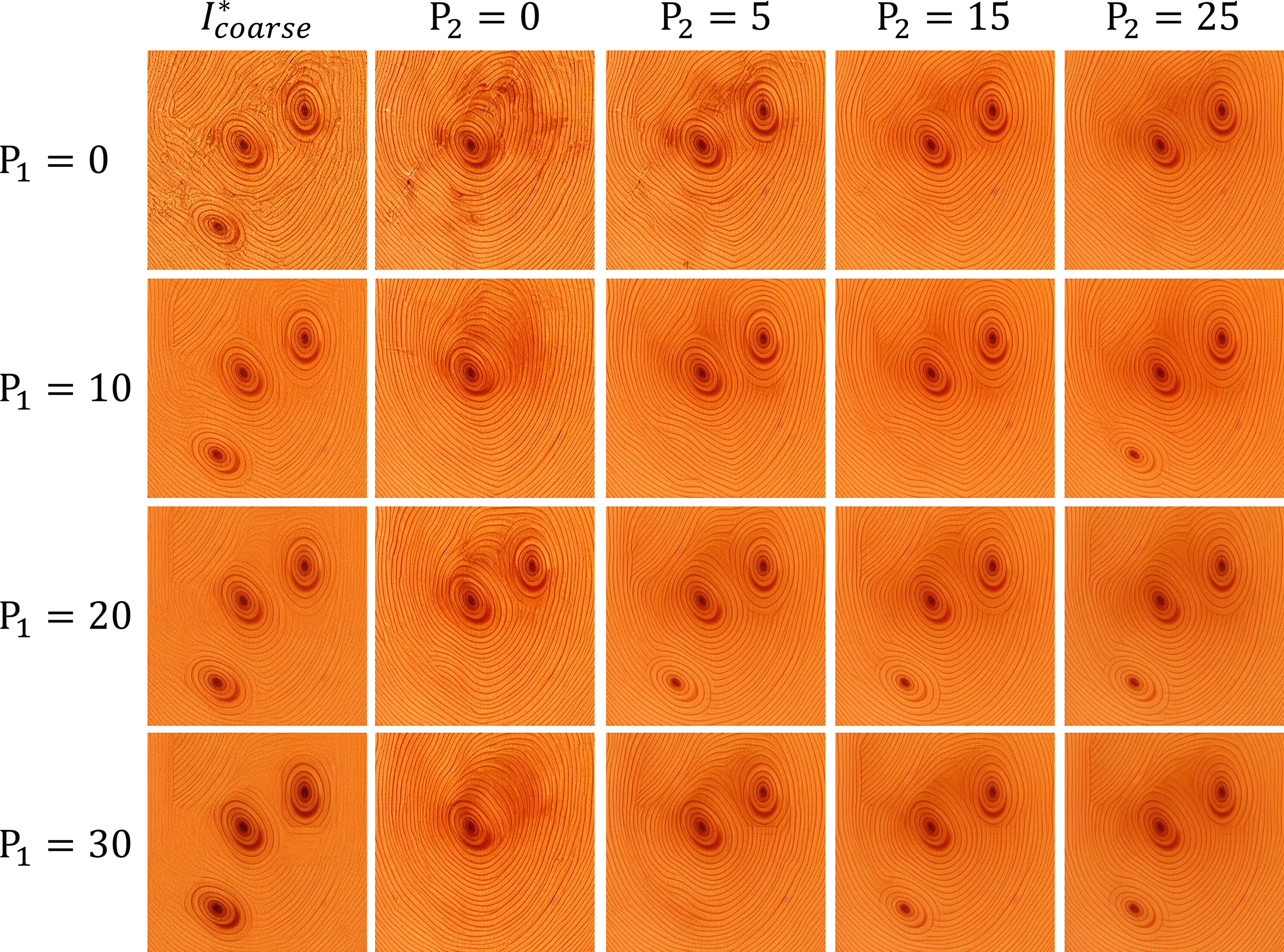}
		\caption{\yang{Ablation study of $KV$-Injection in structure-preserving inversion by exploring the parameter space defined by $\uP_1$ and $\uP_2$}.  Here, we fixed $\uS_1=20$ and $\uS_2=5$ according to the test of Fig.~\ref{fig:s1s2}. While compared with Fig.~\ref{fig:s1s2}, the outputs are significantly improved after introducing $KV$-Injection to the inversion. Many of these results can be considered good synthesis regarding the target image edited and the source texture of the wood.}
		\label{fig:p1p2}
	\end{center}
\end{figure}

\subsection{Ablation Studies}
In this section, we analyze the effect of the key components in our method through ablation studies.

\paragraph{KV-Injection in sampling.} As $KV$-Injection is applied both in the inversion and sampling process, we applied two ablations to study its effect. First, we set $\uP_1=\uP_2=0$, and explore the full \dc{parameter} space for $\uS_1$ and $\uS_2$, to study the effect of $KV$-Injection in texture sampling without being affected by the inversion. 

Fig.~\ref{fig:s1s2} shows a matrix of results of this parameter study. A smaller value of $\uS$ ($\uS_1$ $\&$ $\uS_2$) means performing more time steps of $KV$-Injection in sampling, and thus, the target image is rectified to be closer to the reference texture. On the contrary, a larger S reserves more of the target image's layout, \dc{however, at the same time, it introduces} more structural errors or conflicts, yielding artifacts. None of these results reaches a good trade-off between synthesizing a reasonable global layout (especially considering the user edits) and reproducing local textures of the reference. Nevertheless, we can find a proper parameter setting for S, which is  10$\sim$20 for $\uS_1$, and 5 for $\uS_2$.  Hence we set the default values of $\uS_1$ and $\uS_2$ to be 20 and 5, and use it for all experiments. 

\paragraph{KV-Injection in inversion.} Next, we investigate the effect of $KV$-Injection in our structure-preserving inversion. By fixing $\uS_1$ and $\uS$ to default values, we search the \dc{parameter} space defined by $\uP_1$ and $\uP_2$. As shown in Fig.~\ref{fig:p1p2}, introducing $KV$-Injection significantly improves the synthesis result. Both the structural errors and local artifacts are drastically reduced at the same time. Another important point is that, we may have a relatively wide range for setting the value of $\uP$, which is \dc{empirically} suggested by the results: 10$\sim$30 for $\uP_1$, and 5$\sim$15 for $\uP_2$. \yang{We usually set $\uP_1$ as 20, and $\uP_2$ as 5 in production.} See supplementary for the full exploration results and more examples of this study. 

\begin{figure}[t]
	\begin{center}
		\centering
		\captionsetup{type=figure}
		\includegraphics[width=\linewidth]{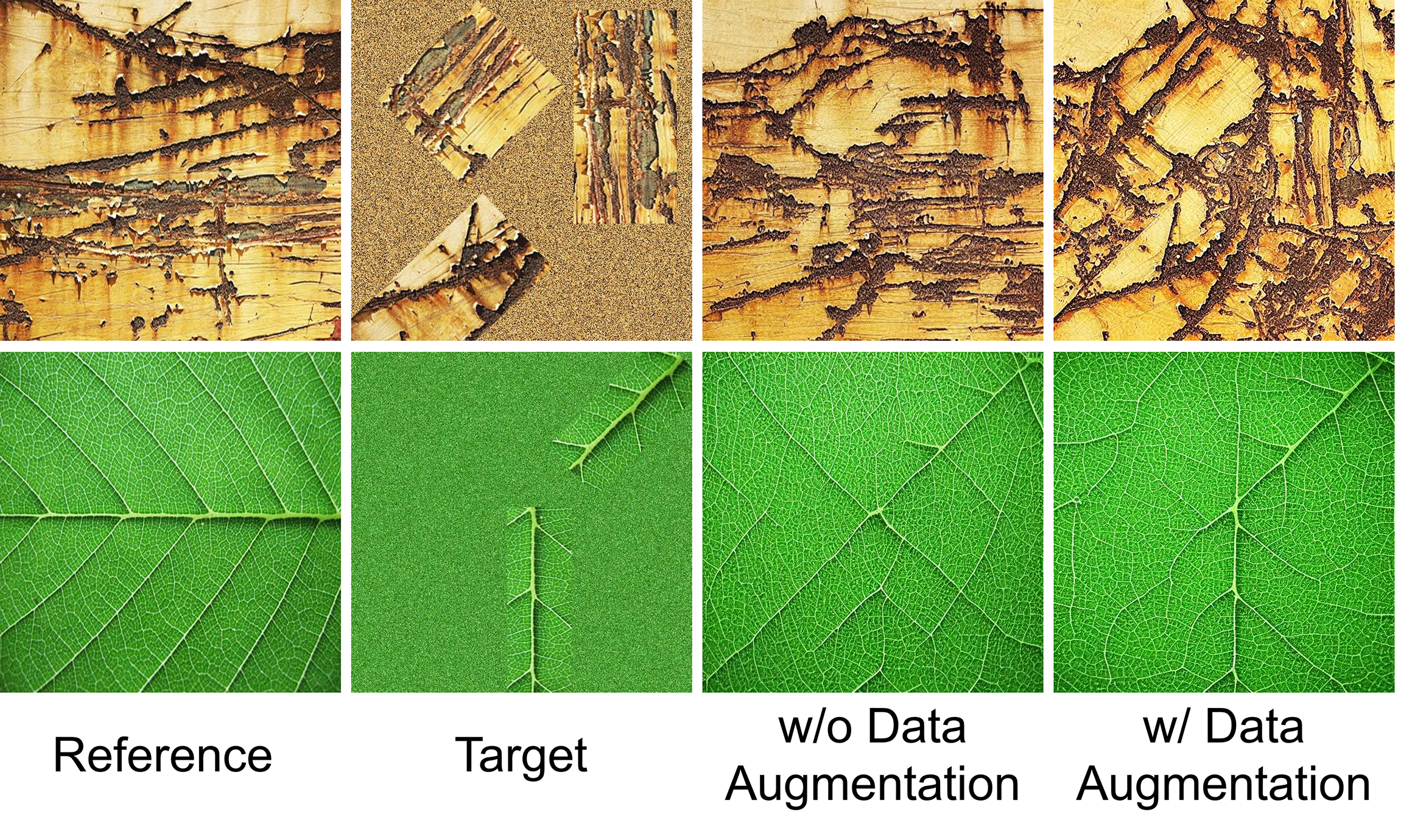}
		\caption{For textures that contain dominant directional structures, data augmentation is essential \dc{to allow the output to admit} with user edits. Specifically, we augment each source shown above with three images rotated at angles of $\pm45^\circ$ and $90^\circ$, respectively.}
		\label{fig:augmentation}
	\end{center}
\end{figure}

\paragraph{Data augmentation.}
In many cases, data augmentation may not be necessary, as the pre-trained Stable Diffusion model \dc{already} has a certain ability to synthesize rotated texture patterns, except when the reference has dominant directional structures. To allow more free editing for directional textures, augmenting the reference is essential to synthesize a more consistent output structure conforming to user edits; see, \eg, \Cref{fig:augmentation} for a comparison on data augmentation, where additional rotated transformations are involved.

\begin{figure}[t]
	\begin{center}
		\centering
		\captionsetup{type=figure}
		\includegraphics[width=\linewidth]{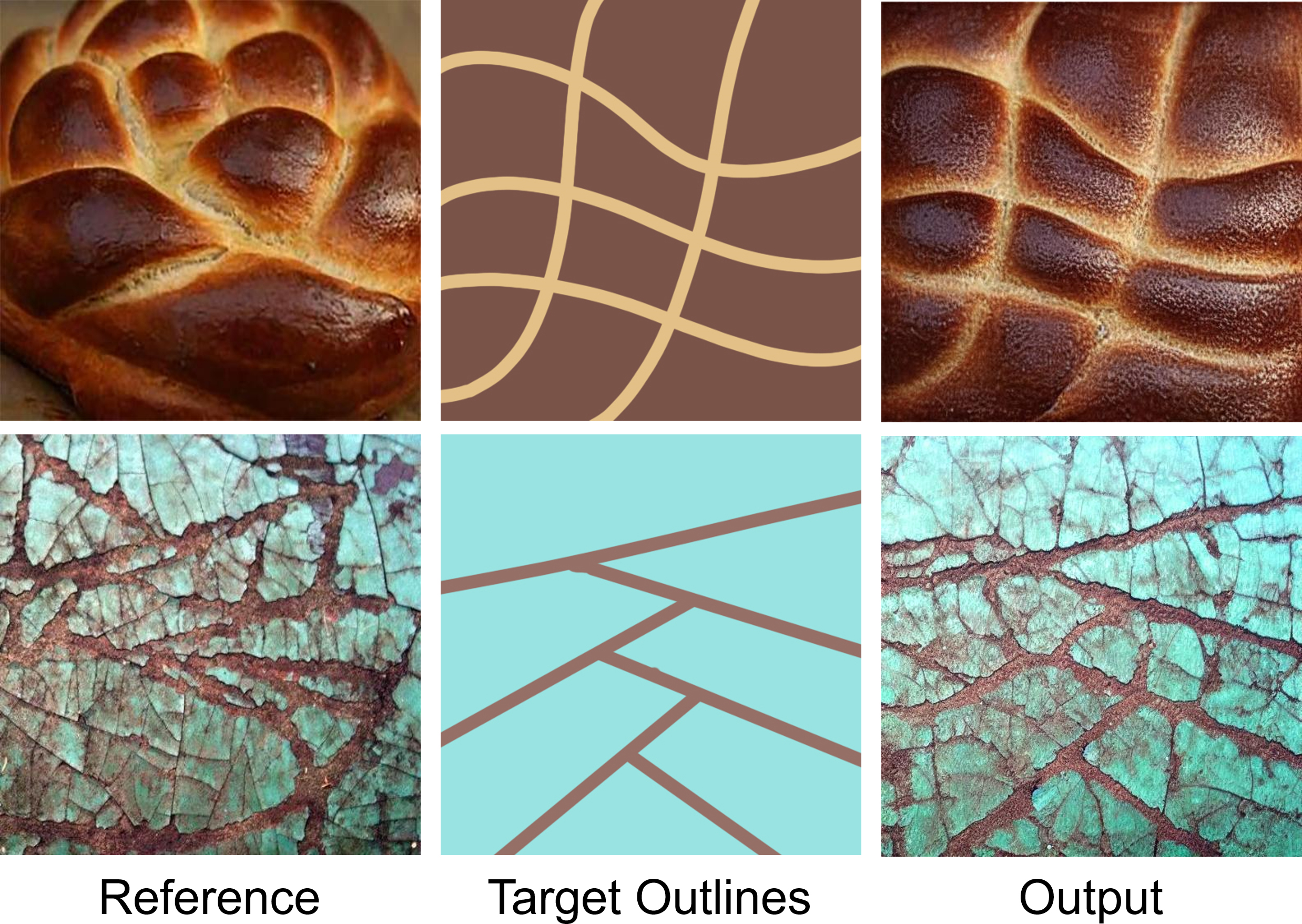}
		\caption{
  Guided texture synthesis. The user provides a color layout to guide the output structure. The synthesized outputs follow the outlines well and still exhibit the reference texture details. 
		}
		\label{fig:transfer}
	\end{center}
\end{figure}

\begin{figure}[t]
	\begin{center}
		\centering
		\captionsetup{type=figure}
		\includegraphics[width=\linewidth]{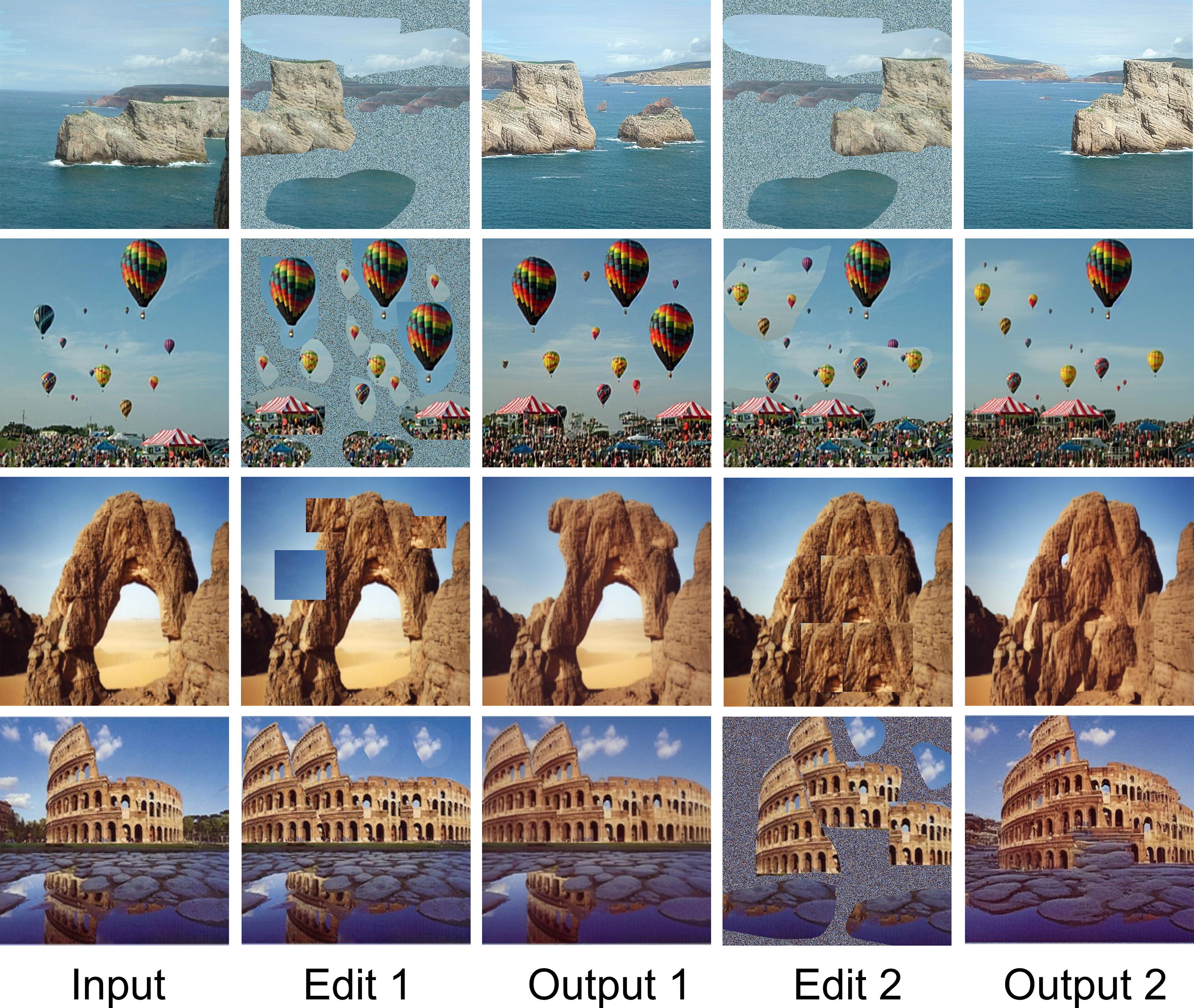}
		\caption{Our method can \dc{also} be  applied to image editing. For \dc{creating} the target image, the user can either \dc{loosely} sketch a layout on the target canvas, or directly edit the input image.}
		\label{fig:imageEditing}
	\end{center}
\end{figure}



\subsection{Additional Applications}
Our self-rectification framework can extend its utility to seemingly other applications. For instance, as illustrated in Fig.~\ref{fig:transfer}, users can input a straightforward target image outlining a layout structure with colors compatible with reference textures. To avoid excessive smoothing, random noise is introduced to enhance the target image. The self-rectification process ensures that the target image incorporates texture details from the reference, while adhering to the user-provided layout structure.

Moreover, our self-rectification method can be employed for image editing beyond textures. Users can edit a given image by employing basic cut-and-paste patch operations, and the resulting crude edits undergo self-rectification. Fig.~\ref{fig:imageEditing} presents several instances of such image editing operations, presenting two edits for each input image.



\section{Conclusions}
\label{sec:future}

Our work addresses the intricate challenge of synthesizing non-stationary textures, offering a method that empowers users to efficiently design new textures with unprecedented controllability. This stands as a notable improvement over existing methods, providing a user-friendly process, which consists of two steps: Users begin with an initial rough edit using conventional image editing tools, followed by an automated self-rectification process. This process leverages a pre-trained diffusion network and injections of self-attention features, showcasing flexibility in synthesizing a diverse range of challenging non-stationary textures.

In the future, we would like to explore the extension of our approach to synthesize large-scale textures. Additionally, there is a promising avenue to incorporate semantic understanding into the self-rectification process, enhancing alignment with user intent. The integration of texture semantics holds the potential to yield contextually relevant and visually appealing synthesized textures.

\paragraph{Acknowledgements}
This work was supported in parts by NSFC (62161146005, U21B2023, U2001206), NSF of Guangdong Province (2022A1515010221), DEGP Innovation Team (2022KCXTD025), Shenzhen Science and Technology Program (KQTD20210811090044003, RCJC20200714114435012), ISF (3441/21), and Guangdong Laboratory of Artificial Intelligence and Digital Economy (SZ).

{\small
\bibliographystyle{ieee_fullname}
\bibliography{egbib}
}

\appendix
\section{Algorithm}

We apply our method on a pre-trained Stable Diffusion (SD) model, which contains an encoder $\mathcal{E}$, a decoder $\mathcal{D}$, and a noise predictor $\epsilon_\theta$. The full pipeline of our method is depicted by ~\Cref{algo:overall,algo:inversion,algo:sampling}. Note the functions Invert(*) and Sample (*) refer to a DDIM inversion step and a DDIM sampling step, respectively, and Att(*) denotes the self-attention mechanism in Stable Diffusion.

\begin{algorithm}
	\caption{Overall Framework}
	\begin{algorithmic}[1]
		\Require Reference texture $\Iref$
		\Ensure Output texture $I^*$
		
		\State $\Itar \gets \text{USER\_EDIT}(\Iref)$
		\State $I^\IR \gets \Itar$
		\State $z^\tar_\T \gets \text{StruPreserving\_Inversion}(\Itar, I^\IR)$
		\State $I^*_\coarse \gets \text{FineTexture\_Sampling}(z_\T^\tar, \Iref)$
		
		\State $z_\T^* \gets \text{StruPreserving\_Inversion}(I^*_\coarse, I^\IR)$ 
		\State $I^* \gets \text{FineTexture\_Sampling}(z_\T^*, \Iref)$
		\State $\textbf{Return}~I^*$
		
	\end{algorithmic}
\label{algo:overall}
\end{algorithm}

\begin{algorithm}
	\caption{Structure-preserving Inversion}
	\begin{algorithmic}[1]
		\Require A target image $\Itar$, an inversion reference $I^\IR$
		\Ensure Inversion code $z_\T^\tar$
		\State $z_0^\IR \gets \mathcal{E}(I^\IR)$
		\State $\{z_0^\IR,z_1^\IR,\ldots,z_\T^\IR\} \gets \text{DDIM\_INVERSION}(z_0^\IR)$
		\State $z_0^\tar \gets \mathcal{E}(\Itar)$
		\For{$t=0, 1, \ldots \uP-1$}
		\State $\{ Q^\IR_{\T-t}, K^\IR_{\T-t}, V^\IR_{\T-t}\} \gets  \epsilon_\theta(z^\IR_{\T-t}, t)$
		\State $\{ Q^\tar_t, K^\tar_t, V^\tar_t\} \gets \epsilon_\theta(z^\tar_{t}, t)$
		\State $\epsilon=\epsilon_\theta(z_t^\tar, t) \sim \text{Att} (Q_t^\tar, K^\IR_{\T-t}, V^\IR_{\T-t}))$ 
		\State $z_{t+1}^\tar \gets \text{Invert}(z_t^\tar, \epsilon, t)$ 
		\EndFor
		\For{$t=\uP, \uP+1, \ldots, \T-1$}
		\State $\{ Q^\tar_t, K^\tar_t, V^\tar_t\} \gets \epsilon_\theta(z^\tar_{t}, t)$
		\State $\epsilon=\epsilon_\theta(z_t^\tar, t) \sim \text{Att}(Q_t^\tar, K^\tar_t, V^\tar_t)$ 
		\State $z_{t+1}^\tar \gets \text{Invert}(z_t^\tar, \epsilon, t)$  
		\EndFor
		\State $\textbf{Return}~z^\tar_\T$
	\end{algorithmic}
	\label{algo:inversion}
\end{algorithm}

\begin{algorithm}
	\caption{Fine-texture Sampling}
	\begin{algorithmic}[1]
		\Require A start code $z^*_\T$, a reference texture $\Iref$
		\Ensure Output texture $I^*$
		
		\State $z_0^\R \gets \mathcal{E}(\Iref)$
		\State $\{z_0^{\R},z_1^{\R},\ldots,z_\T^{\R}\} \gets \text{DDIM\_INVERSION}(z_0^\R)$
		\For{$t=\T, \T-1, \ldots, \T-\uS-1$}
		\State $\{ Q^{*}_t, K^{*}_t, V^{*}_t\} \gets \epsilon_\theta(z_t^*, t)$
		\State $\epsilon = \epsilon_\theta(z_t^*, t) \sim \text{Att}(Q^{*}_t, K^{*}_t, V^{*}_t)$
		\State $z_{t-1}^* \gets \text{Sample}(z_t^*, \epsilon, t)$
		\EndFor
		
		\For{$t=\T-\uS, \T-\uS+1, \ldots, 1$}
		\State $\{ Q^{\R}_t, K^{\R}_t, V^{\R}_t\} \gets \epsilon_\theta(z_t^\R, t)$
		\State $\{ Q^{*}_t, K^{*}_t, V^{*}_t\} \gets \epsilon_\theta(z_t^*, t)$
		\State $\epsilon = \epsilon_\theta(z_t^*, t) \sim \text{Att}(Q^{*}_t, K^{\R}_t, V^{\R}_t)$
		\State $z_{t-1}^* \gets \text{Sample}(z_t^*, \epsilon, t)$
		\EndFor
		\State $I^* \gets \mathcal{D}(z_0^*)$
		\State $\textbf{Return}~I^*$
	\end{algorithmic}
	\label{algo:sampling}
\end{algorithm}

\end{document}